\documentclass{article}

\usepackage[preprint]{neurips_2024}
\pdfoutput=1

\usepackage{natbib}
\setcitestyle{numbers,square,comma}%

\usepackage{multirow}
\usepackage{colortbl}
\usepackage{makecell}
\usepackage{arydshln}
\usepackage{booktabs}
\usepackage{bm}
\usepackage{amsmath}
\usepackage{marvosym}
\usepackage{graphicx}
\usepackage{subcaption}
\usepackage{wrapfig}

\usepackage{amsfonts}
\usepackage{graphicx}
\definecolor{cvprblue}{rgb}{0.21,0.49,0.74}
\usepackage[pagebackref,breaklinks,colorlinks,allcolors=cvprblue]{hyperref}
\usepackage{graphicx}

\title{TinyLLaVA-Video-R1: Towards Smaller LMMs for Video Reasoning}

\author{\quad Xingjian Zhang$^{1,}$\thanks{denotes~ equal ~contributor; ~\textrm{\Letter}~denotes corresponding author.} \quad Siwei Wen$^{1,2,\ast}$ \quad Wenjun Wu$^{1,2,3}$ \quad  Lei Huang$^{1,2,3,~\textrm{\Letter}}$
\\
\\
{\small$^{1}$SKLCCSE, Institute of Artificial Intelligence, Beihang University, Beijing, China}\\
{\small $^{2}$Beijing Advanced Innovation Center for Future Blockchain and Privacy Computing, Beihang University}\\
{\small $^{3}$Hangzhou International Innovation Institute, Beihang University, Hangzhou, China}\\
\\
\normalsize\texttt{\{huangleiai\}@buaa.edu.cn}
}

\begin{document}
\maketitle
\begin{abstract}

Recently, improving the reasoning ability of large multimodal models (LMMs) through reinforcement learning has made great progress. However, most existing works are based on highly reasoning-intensive datasets such as mathematics and code, and researchers generally choose large-scale models as the foundation. 
We argue that exploring small-scale models' reasoning capabilities remains valuable for researchers with limited computational resources. Moreover, enabling models to explain their reasoning processes on general question-answering datasets is equally meaningful. Therefore, we present the small-scale video reasoning model TinyLLaVA-Video-R1. Based on TinyLLaVA-Video \cite{zhang2025tinyllava}, a traceably trained video understanding model with no more than 4B parameters, it not only demonstrates significantly improved reasoning and thinking capabilities after using reinforcement learning on general Video-QA datasets, but also exhibits the emergent characteristic of "aha moments". Furthermore, we share a series of experimental findings, aiming to provide practical insights for future exploration of video reasoning (thinking) abilities in small-scale models. It is available at \url{https://github.com/ZhangXJ199/TinyLLaVA-Video-R1}.

\end{abstract}
\section{Introduction}
\label{sec:intro}

Since DeepSeek-R1 \cite{guo2025deepseek} demonstrated that pure reinforcement learning can significantly enhance a model's reasoning capabilities, many subsequent works \cite{chen2025r1v, huang2025vision, meng2025mm, zhou2025r1, peng2025lmm} have also explored improving the reasoning abilities of multimodal models, achieving notable progress. Most of these efforts focus on extending reasoning capabilities to the image modality \cite{meng2025mm, chen2025r1v}, conducting research using strong reasoning data such as math-image pairs and spatial reasoning \cite{peng2025lmm, huang2025vision, zhou2025r1}, or task-specific data like grounding \cite{liu2025visual, shen2025vlmr1}. However, existing research on video reasoning models has not made significant progress due to the scarcity of highly reasoning-intensive data.


Open-R1-Video \cite{wang-2025-open-r1-video} is the first to introduce reasoning into the video domain, however, its performance on general Video-QA datasets is unsatisfactory, with benchmark results even showing a decline. Subsequently, Video-R1 \cite{feng2025video} successfully integrates strong reasoning image-text pairs for video reasoning, achieving remarkable performance, but they make preliminary attempts and argue that small-scale models cannot produce effective reasoning processes. However, the high computational cost of large-scale models remains a significant barrier for many researchers with limited resources. Therefore, exploring the reasoning capabilities of smaller models is still necessary. 


In this work, we propose the small-scale video reasoning model TinyLLaVA-Video-R1, based on the traceably trained model TinyLLaVA-Video \cite{zhang2025tinyllava}. After reinforcement learning on general Video-QA datasets, the model not only significantly improves its reasoning and thinking abilities, but also exhibits the emergent characteristic of “aha moments”, which is more meaningful than simply generating answers through perception. Moreover, through extensive experiments under various configurations, we have obtained a series of insightful findings. We believe these discoveries will provide valuable guidance for future exploration of video reasoning capabilities in small-scale models.


\begin{figure*}[t]
  \centering
  \includegraphics[width=\textwidth]{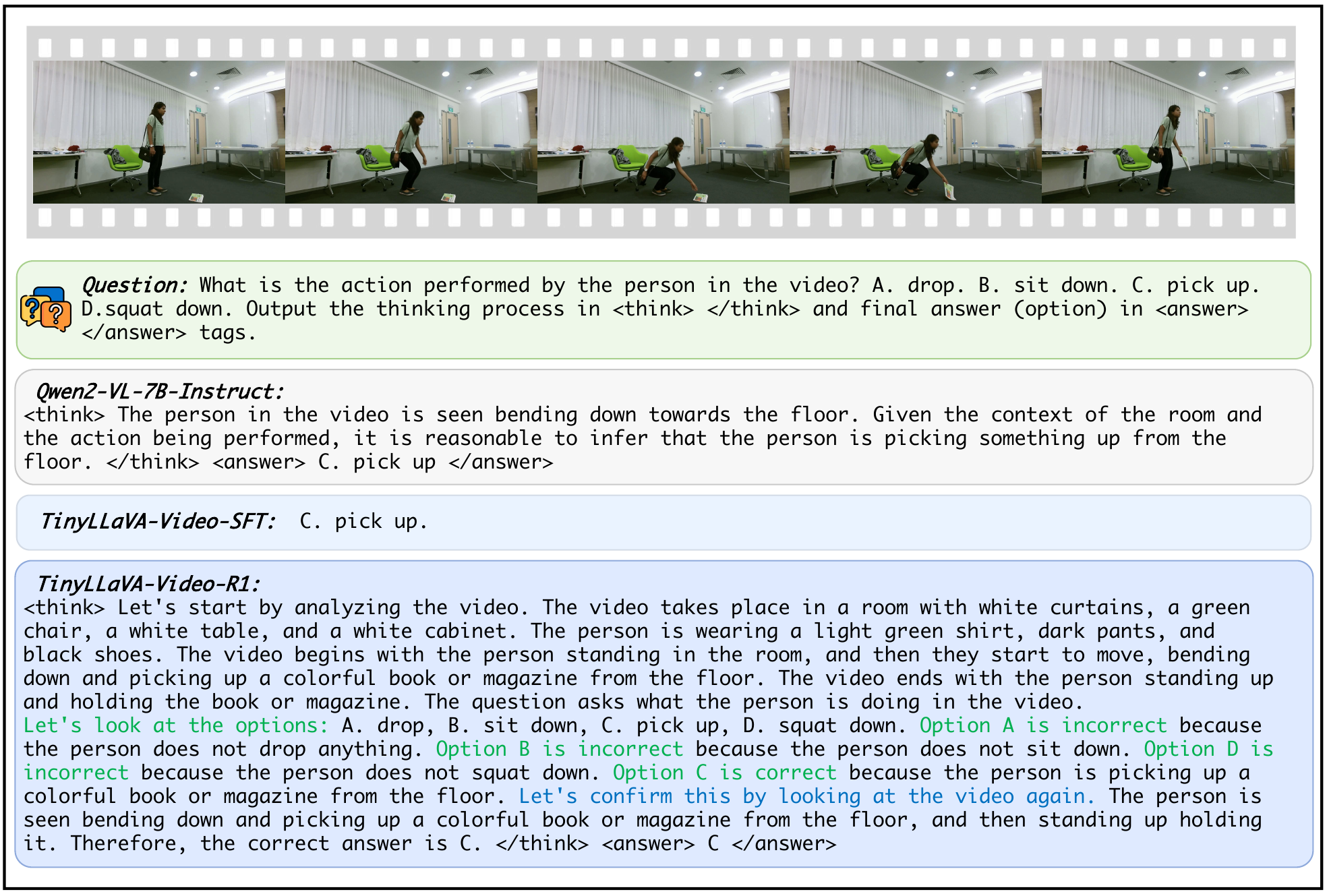}
   \caption{A case of TinyLLaVA-Video-R1 on video understanding data, sourced from MVBench. The model demonstrates the ability to perceive video scenes and analyze options, while also exhibiting reflective and backtracking behavior (highlighted in blue).}
   \label{fig:case1}
   \vspace{-0.1in}
\end{figure*}
\section{Related Work}
\label{sec:formatting}

\paragraph{Large Language Reasoning Models.}
Recently, Kimi K1.5 \cite{team2025kimi} and DeepSeek-R1 \cite{guo2025deepseek} have gained widespread attention for their significant improvements in model performance on reasoning tasks through using reinforcement learning. Unlike approaches relying on process-supervised reward models \cite{guan2025rstar}, the Group Relative Policy Optimization (GRPO) algorithm \cite{shao2024deepseekmath} relying on rule-based rewards not only substantially reduces computational costs but has also sparked a new wave of interest due to intriguing phenomena such as "aha moments" observed during training. Under the influence of this research paradigm, numerous follow-up studies have advanced along this technical path, some have successfully replicated the method and extended it to multimodal domains \cite{chen2025r1v, huang2025vision, zhou2025r1}, achieving notable progress, while others \cite{liu2025understanding, yu2025dapo} have focused on algorithmic optimization to further enhance the reasoning efficiency and performance of models.

\paragraph{Multimodal Reasoning Models.} 

With the remarkable advancement in reasoning capabilities of language models, an increasing number of studies have attempted to extend them to multimodal domains. LMM-R1 \cite{peng2025lmm} proposes a two-stage training strategy to enhance the reasoning performance of multimodal models; Vision-R1 \cite{huang2025vision} attempts to address the post-cold-start overthinking issue in multimodal models; Video-R1 \cite{feng2025video} develops T-GRPO to further enhance model's video comprehension. While these studies have made notable progress, most tend to adopt base models with 7B or more parameters to ensure superior reasoning effectiveness and robust performance.

Although some preliminary work has explored small-scale multimodal models \cite{peng2025lmm, chen2025r1v, zhou2025r1}, these investigations have primarily focused on the image modality, leaving research on video reasoning significantly underdeveloped. Due to the scarcity of high-quality video reasoning datasets and the inherent capability limitations of models, small-scale video reasoning models face multiple challenges including limited response length, difficulties in presenting coherent reasoning chains, and constrained overall performance improvement \cite{feng2025video}. These critical challenges demand thorough investigation and substantive technical breakthroughs.

\begin{figure*}[t]
  \centering
  \includegraphics[width=\textwidth]{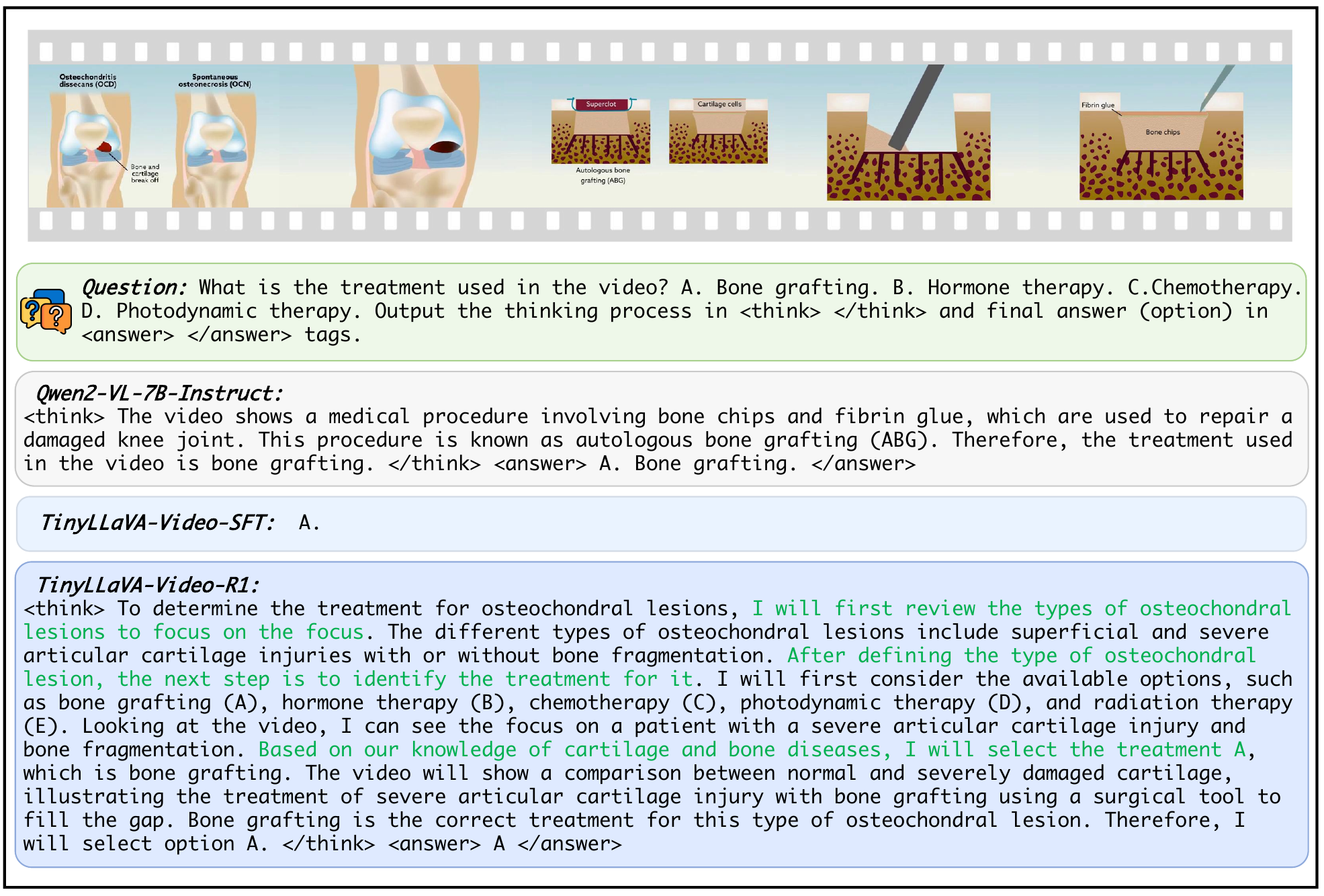}
   \caption{A case of TinyLLaVA-Video-R1 on video reasoning data, sourced from MMVU. The model demonstrates comprehensive video content understanding and the capability to derive correct answers through analytical reasoning.}
   \label{fig:case2}
   \vspace{-0.15in}
\end{figure*}

\vspace{-0.05in}
\section{Methods}
\vspace{-0.05in}

To explore the video reasoning capabilities of small-scale models, we conduct experiments on TinyLLaVA-Video \cite{zhang2025tinyllava}. We utilize the GRPO algorithm on the general Video-QA dataset NextQA and made specific modifications to the reward rules: adding a continuous length reward to the format reward and introducing penalties for incorrect answers. The experimental results in Section \ref{sec:experiments} demonstrate the effectiveness of these modifications.

\vspace{-0.05in}
\subsection{TinyLLaVA-Video}

TinyLLaVA-Video is a fully open-source small-scale video understanding model that employs Qwen2.5-3B \cite{hui2024qwen2} as its language model and SigLIP \cite{zhai2023sigmoid} as its visual encoder. It delivers competitive performance across multiple benchmarks. Crucially, its training data are fully open-sourced, and the entire training process remains traceable. This effectively prevents the repeated use of identical data across different training phases, thereby avoiding the introduction of uncontrolled variables and ensuring more reliable experimental results and conclusions. Such reproducibility and controllability represent a distinct advantage over models that only release weights, making TinyLLaVA-Video an ideal foundational model for our experiments on investigating video reasoning.

\begin{figure*}[t]
  \centering
  \includegraphics[width=\textwidth]{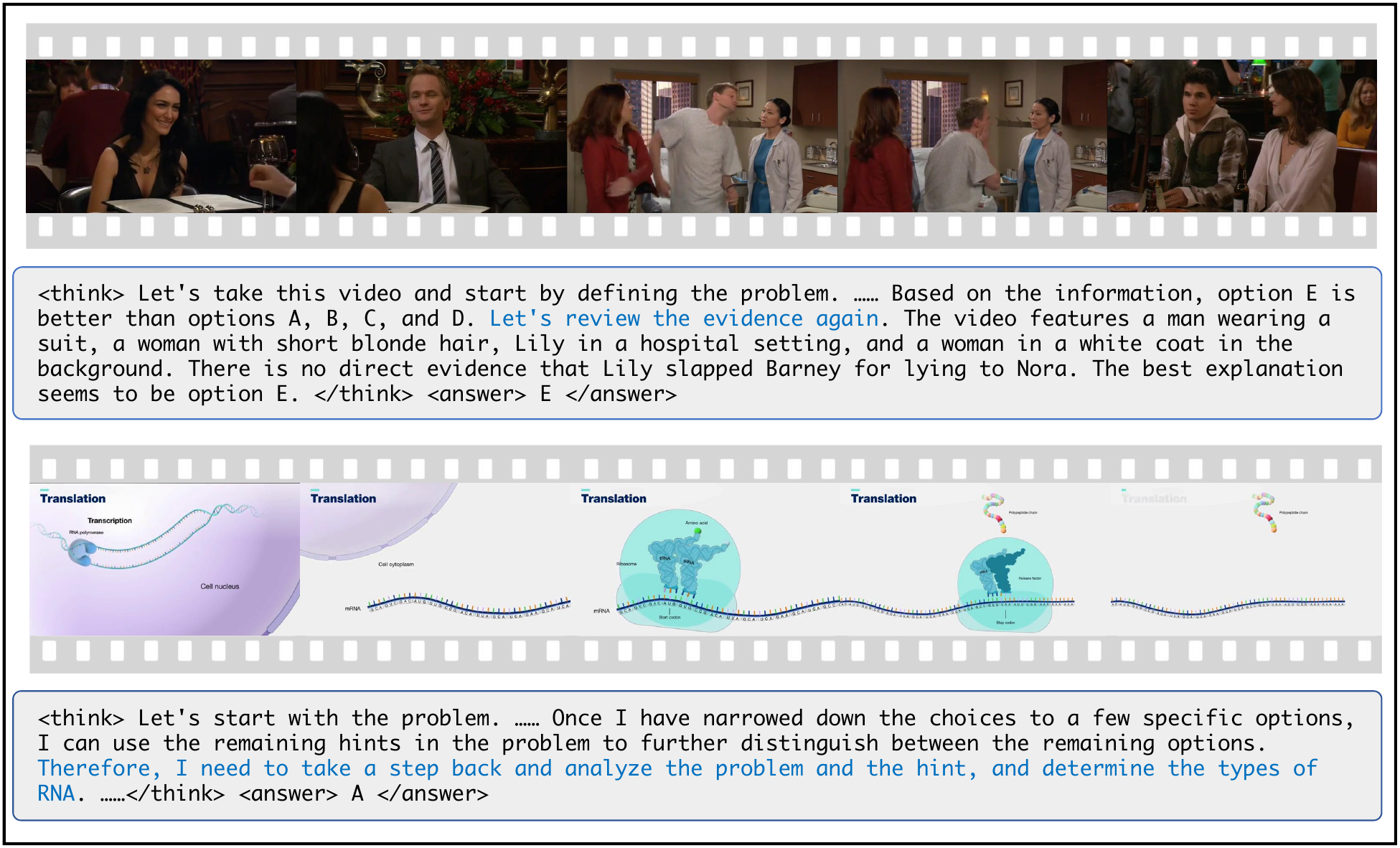}
   \vspace{-0.15in}
   \caption{Cases of "aha moment", where the model demonstrates reflection and backtracking during its reasoning process (highlighted in blue). The cases are from MVBench and MMVU respectively.}
   \label{fig:aha}
   \vspace{-0.1in}
\end{figure*}

\vspace{-0.05in}
\subsection{Group Relative Policy Optimization (GRPO)}

We follow the GRPO algorithm \cite{shao2024deepseekmath} to train the model. For each question q, the policy model generates a set of candidate responses \(\{{O}_1, {O}_2, \dots, {O}_G \}\), computes the corresponding rewards \(\{{r}_1, {r}_2, \dots, {r}_G \}\) based on the reward rules. And then these rewards are normalized to calculate the advantage for each response. Subsequently, the model is optimized through maximization of the following objective function:
\begin{equation}
\resizebox{0.9\textwidth}{!}{ 
    $J_{GRPO}(\theta) = \mathbb{E}_{[q,\{o_i\}]} \frac{1}{G}\sum_{i=1}^G\frac{1}{|o_i|} \bigg\{ \min\left[ \frac{\pi_{\theta}}{\pi_{\theta_{old}}}A_{i}, \right.
    \left. \text{clip}\left(\frac{\pi_{\theta}}{\pi_{\theta_{old}}}, 1-\epsilon, 1+\epsilon\right) A_{i} \right] - \beta \mathbb{D}_{KL}\left[\pi_{\theta} \|\pi_{ref}\right] \bigg\}$
}
\label{eq:1}
\end{equation}
where $\pi_{\theta}$ and $\pi_{\theta_{old}}$ are the current and old policy, $\epsilon$ and $\beta$ are hyper-parameters, and $A_i$ is the advantages defined as: 
\begin{equation}
A_i = \frac{r_i - \text{mean}(\{r_1, r_2, \cdots, r_G\})}{\text{std}(\{r_1, r_2, \cdots, r_G\})}.
\label{eq:2}
\end{equation}

In addition, our experimental observations reveal an issue analogous to DAPO \cite{yu2025dapo}: when all responses in a set $\{{O}_i\}$ are correct and given equal rewards, their computed advantages vanish to zero. This phenomenon affects policy updates and diminishes sample efficiency. To maximize the utility of each sample, we introduce an additional gaussian noise $\mathcal{N}(0,\,0.02^2)$ to the advantages. Although the noise induces only minor perturbations, it ensures intra-group advantage diversity across responses. 

\subsection{Training Data and Template.} 

We select multiple choice questions from the NextQA \cite{xiao2021next} subset of LLaVA-Video-178K \cite{zhang2024video} as training data. To maintain manageable training time with limited computational resources, we only choose the subset of data with a duration of 0 to 30 seconds, which contains 5,496 samples. It is a weak reasoning dataset, where the questions are more perception-oriented and exhibit weaker logical reasoning. However, we hypothesize that the model's reasoning abilities are likely predominantly derived from reinforcement learning, and we still aim to guide it to demonstrate its thought process by articulating the reasoning behind its choices, rather than merely providing an answer.

During training, for each input question, in addition to the system template, we append the following prompt at the end: 
\texttt{Output the thinking process in <think> </think> and final answer (option) in <answer> </answer> tags.} 
Moreover, when computing rewards for responses, we strictly enforce the model to adhere to this format.

\subsection{Reward Rules.} 
\label{reward rules}
We also avoid using a reward model and define reward rules based on the format and accuracy of the responses as follows:

\paragraph{Format reward.} 
We require the thought process to be enclosed within \texttt{<think></think>}, and the final answer to be enclosed within \texttt{<answer></answer>}. These four tags can appear only once in the entire response, and if followed, the model will receive a format reward $FR = r_0 + LR$. 
Here, $r_0$ represents the base reward for adhering to the required response format, and $LR$ is the continuous length reward designed to encourage the model to generate longer outputs, calculated as: 
\begin{equation}
LR = \min\left(1, \frac{Len}{ML}\right) \times r_1 .
\label{eq:3}
\end{equation}
Here, $Len$ represents the length of the response extracted from within the \texttt{<think></think>} tags, and $ML$ represents the maximum length corresponding to the upper limit of the reward. In our experiments, we set $r_0 = r_1 = 0.5$, thus the format reward is limited to a maximum of 1.

\paragraph{Accuracy reward.} We design the accuracy reward $AR$ based on the answer. We extract the final answer from \texttt{<answer></answer>} and compare it with the label. The model will receive an accuracy reward of $AR = r_2 > 0$, if the answer is correct. Responses with either format errors preventing answer extraction or incorrect answers will result in zero accuracy reward, i.e. $AR = 0$. To ensure that the accuracy reward and the format reward have equal importance, we set $r_2 = r_0 + r_1 $ in our experiments. \\

To encourage the model to increase the response length only when answering correctly, rather than arbitrarily increasing the length at the cost of accuracy, we deviate from most existing approaches that simply define the total reward as the sum of format reward and accuracy reward. Instead, we introduce a penalty for incorrect answers, with the total reward $R$ defined by the following formula:
\begin{equation}
R = 
\begin{cases} 
AR + FR, & \text{if } FR > 0 \text{ and } AR = r_2 \\ 
- FR, & \text{if } FR > 0 \text{ and } AR = 0 \\ 
- ( r_0 + r_1 + r_2), & \text{if } FR = 0 
\end{cases}
\label{eq:4}
\end{equation}
When the model's answer is correct, the longer the reasoning process, the higher the reward. In contrast, if the answer is incorrect, the longer the reasoning process, the higher the penalty incurred.

\begin{table}[t]
\centering
\footnotesize
\setlength{\tabcolsep}{1.5pt} 
\renewcommand{\arraystretch}{1.3} 
\begin{tabular*}{\textwidth}{@{}c@{\extracolsep{\fill}}cccccc@{}}
    \toprule
    Model & LLM size & Answer Type & MVBench & Video-MME (wo sub) & MLVU & MMVU (mc) \\
    \midrule
    LLaMA-VID \cite{li2025llama} & 7B & Option & 41.4 & - & 33.2 & - \\
    LLaVA-NeXT \cite{liu2024llava} & 7B & Option & - & - & 39.3 & 29.2 \\
    VideoLLaVA \cite{lin2023video} & 7B & Option & - & 39.9 & 47.3 & - \\
    ShareGPT4Video \cite{chen2024sharegpt4video} & 8B & Option & - & 39.9 & 46.4 & - \\
    LLaVA-Mini \cite{zhang2025llava} & 7B & Option & 44.5 & - & 42.8 & - \\
    InternVideo2 \cite{wang2024internvideo2} & 8B & Option & - & 41.9 & - & 39.0 \\
    \midrule
    TinyLLaVA-Video-SFT & 3B & Option & 49.0 & 42.2 & 49.2 & 46.1 \\
    TinyLLaVA-Video-ColdStart & 3B & Reason & 33.2 & 26.6 & 28.6 & 22.7 \\
    \midrule
    \textbf{TinyLLaVA-Video-R1} & 3B & Reason & \textbf{49.5} & \textbf{46.6} & \textbf{52.4} & \textbf{46.9} \\
    \bottomrule
\end{tabular*}
\vspace{3mm}
\caption{The performance of TinyLLaVA-Video-R1 on multiple benchmarks. "Option" indicates that the model only needs to answer with the selected choice, while "Reason" means the model must output both the answer and the reasoning process according to the format requirements. Here, MMVU is categorized as a video reasoning benchmark, the remaining benchmarks are designed for general-purpose video evaluation. The best results are indicated by \textbf{boldface}.}
\label{tab:performance}
\vspace{-0.2in}
\end{table}

\section{Experiments}
\label{sec:experiments}

\subsection{Experimental Settings}

We conduct experiments on 8 NVIDIA A100-40G GPUs. During training, we keep the vision encoder frozen and update the connector and language model. We set the learning rate at 1e-6 for stable training. 

To facilitate rapid adaptation to reasoning format and ensure training stability, we first finetune the model using 16 human-annotated cold-start samples, resulting in TinyLLaVA-Video-ColdStart. We then adopt it as the base model for reinforcement learning and train on 5,496 NextQA data for one epoch to obtain TinyLLaVA-Video-R1.

For evaluation, we select four commonly used video understanding and reasoning benchmarks: MVBench \cite{li2024mvbench}, VideoMME\cite{fu2024video}, MLVU \cite{zhou2024mlvu}, and MMVU \cite{zhao2025mmvu}. These benchmarks encompass videos from multiple disciplines and domains, with a wide range of durations, enabling a comprehensive assessment of the model's capabilities.

\subsection{Main Results and Aha Moment}

\begin{figure}[h!]
  \centering
  \begin{subfigure}[b]{0.32\textwidth}
    \includegraphics[width=\linewidth]{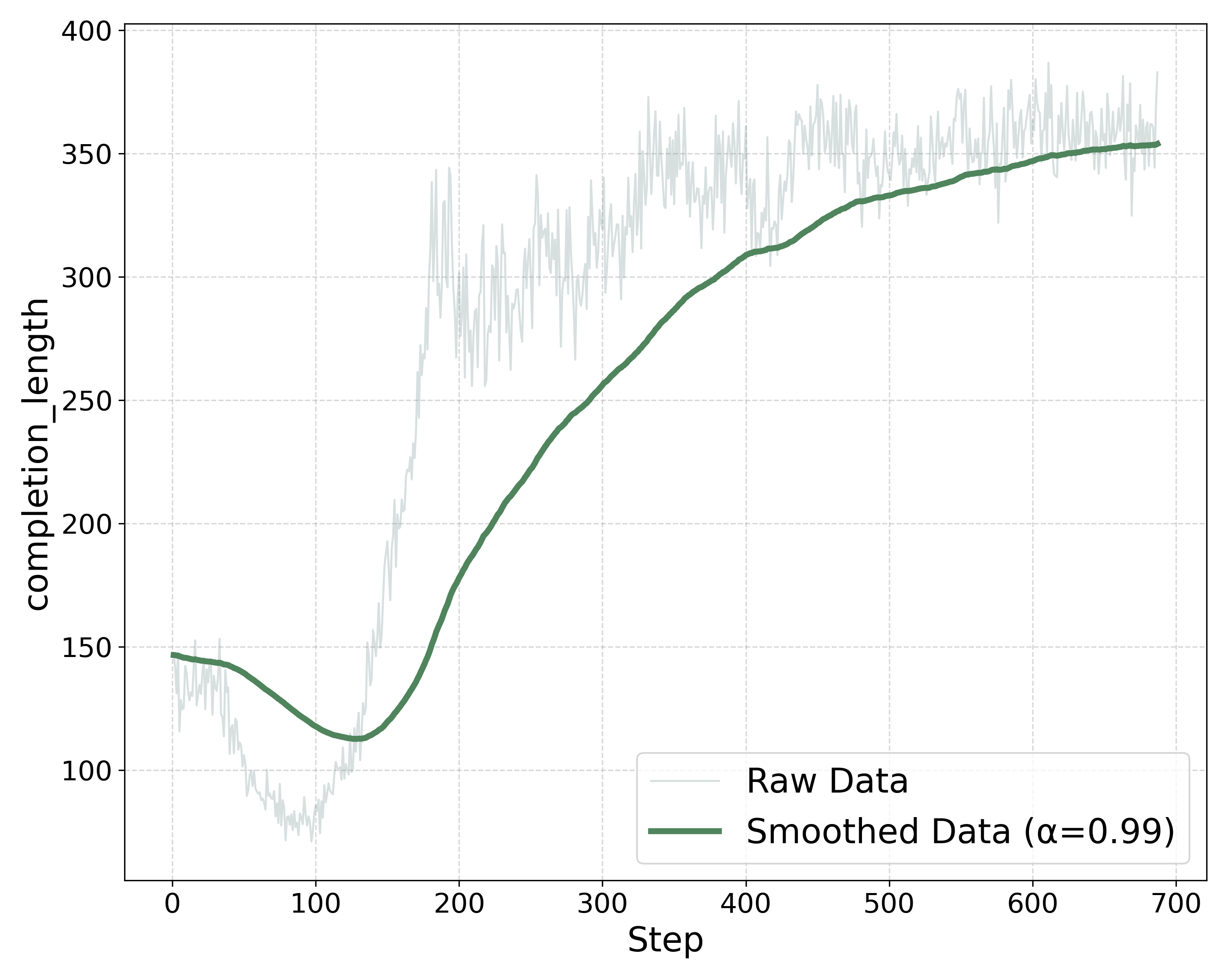}
    \caption{\small Evolution in completion length.}
    \label{fig:short-a}
  \end{subfigure}
  \hfill
  \begin{subfigure}[b]{0.32\textwidth}
    \includegraphics[width=\linewidth]{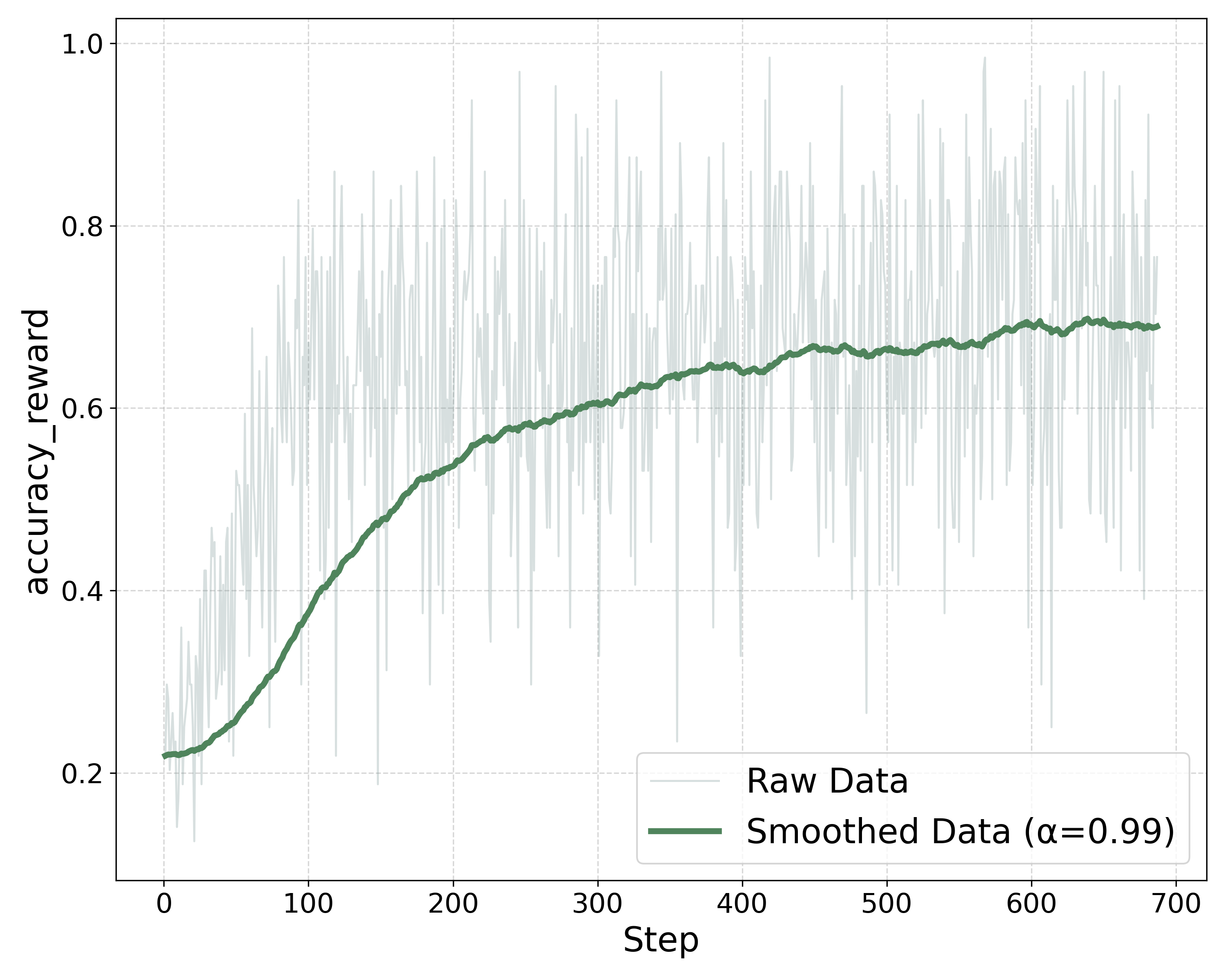}
    \caption{\small Evolution in accuracy reward.}
    \label{fig:short-b}
  \end{subfigure}
  \hfill
  \begin{subfigure}[b]{0.32\textwidth}
    \includegraphics[width=\linewidth]{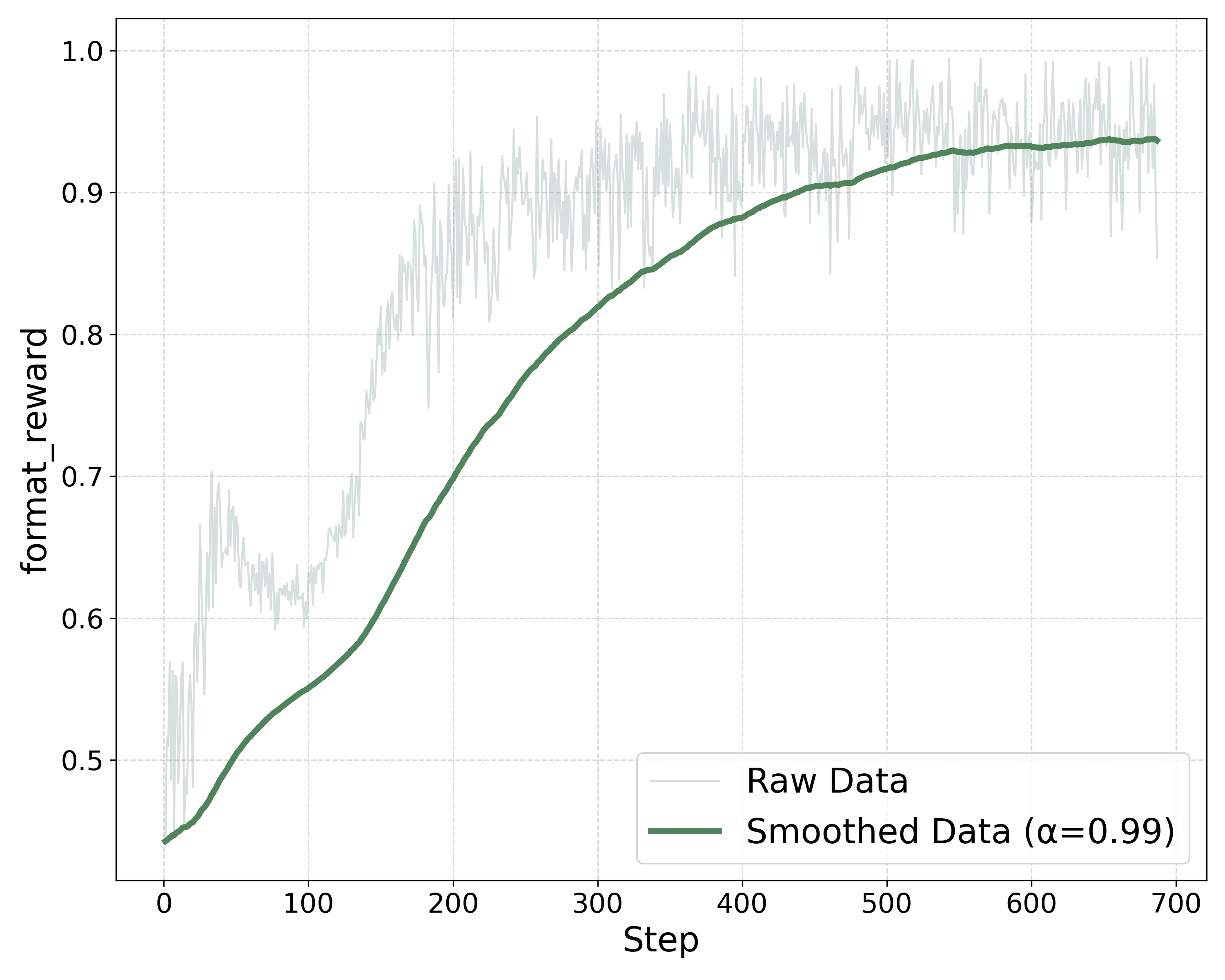}
    \caption{\small Evolution in format reward.}
    \label{fig:short-c}
  \end{subfigure}
  \caption{Evolution in key metrics during the training of TinyLLaVA-Video-R1. Under our reward rule settings, both the response length and rewards of TinyLLaVA-Video-R1 gradually increased during training.}
  \label{fig:short}
\end{figure}

As shown in Figure \ref{fig:short}, during training, both the response length and rewards demonstrate stable growth. As presented in Table \ref{tab:performance}, compared to TinyLLaVA-Video-SFT, which is trained on the same dataset using supervised learning, TinyLLaVA-Video-R1 shows superior performance across multiple benchmarks. Additionally, compared to the base model TinyLLaVA-Video-ColdStart, TinyLLaVA-Video-R1 not only adheres to the required response format but also demonstrates improved reasoning capabilities. 

As shown in Figures \ref{fig:case1} and \ref{fig:case2}, we further illustrate the model's reasoning ability. The model can comprehend and analyze video content, evaluate each option step-by-step, and ultimately provide an answer. Compared to models that only output answers without reasoning, TinyLLaVA-Video-R1 generates meaningful thought processes, making its responses more interpretable and valuable. This represents a significant advantage of video reasoning models over conventional video understanding models.

Similar to other works that use reinforcement learning to enhance model reasoning capabilities, we also reproduce the "aha moment" in TinyLLaVA-Video-R1, where the model exhibits emergent behaviors such as self-verification during its reasoning process. Our experimental results confirm that even when trained with weakly-reasoned general video data through reinforcement learning, the smaller model can still demonstrate retrospection and reflection. 

As highlighted in the blue annotations in Figures \ref{fig:case1} and \ref{fig:aha}, the model revisits and verifies its initial reasoning after completing a round of thought. This behavior indicates that the model does not merely perform perception but also engages in continuous thinking and self-checking.

\subsection{Ablation Study}
In this section, we present ablation studies on methods and key experimental findings that contribute significantly to the performance enhancement of TinyLLaVA-Video-R1.

\subsubsection{Impact of Cold-Start Data}
Due to the limitations of language models, when we directly use TinyLLaVA-Video as the base model without length reward, we find that as training progresses, the model has a certain probability of learning to 'take shortcuts'. While adhering to the required format, all responses omit the reasoning process and are structured strictly as \texttt{<think> </think> <answer> option </answer>}. We observe similar experimental phenomena when conducting experiments on Qwen2-VL-2B \cite{wang2024qwen2}, so we believe this is a common issue with small-scale models. 

However, when we perform a cold start with 16 human-annotated CoT data, this phenomenon no longer appear during the experiments. At the same time, the model also learn to comply with the format requirements more quickly. Therefore, we believe that cold starting is necessary for reasoning in small-scale models. Even a small amount of cold start data can be very helpful for stabilizing model training.

\subsubsection{Impact of Refinement of Format Rewards}

\begin{wrapfigure}{r}{0.45\textwidth} 
  \vspace{-0.15in}
  \centering
  \includegraphics[width=\linewidth]{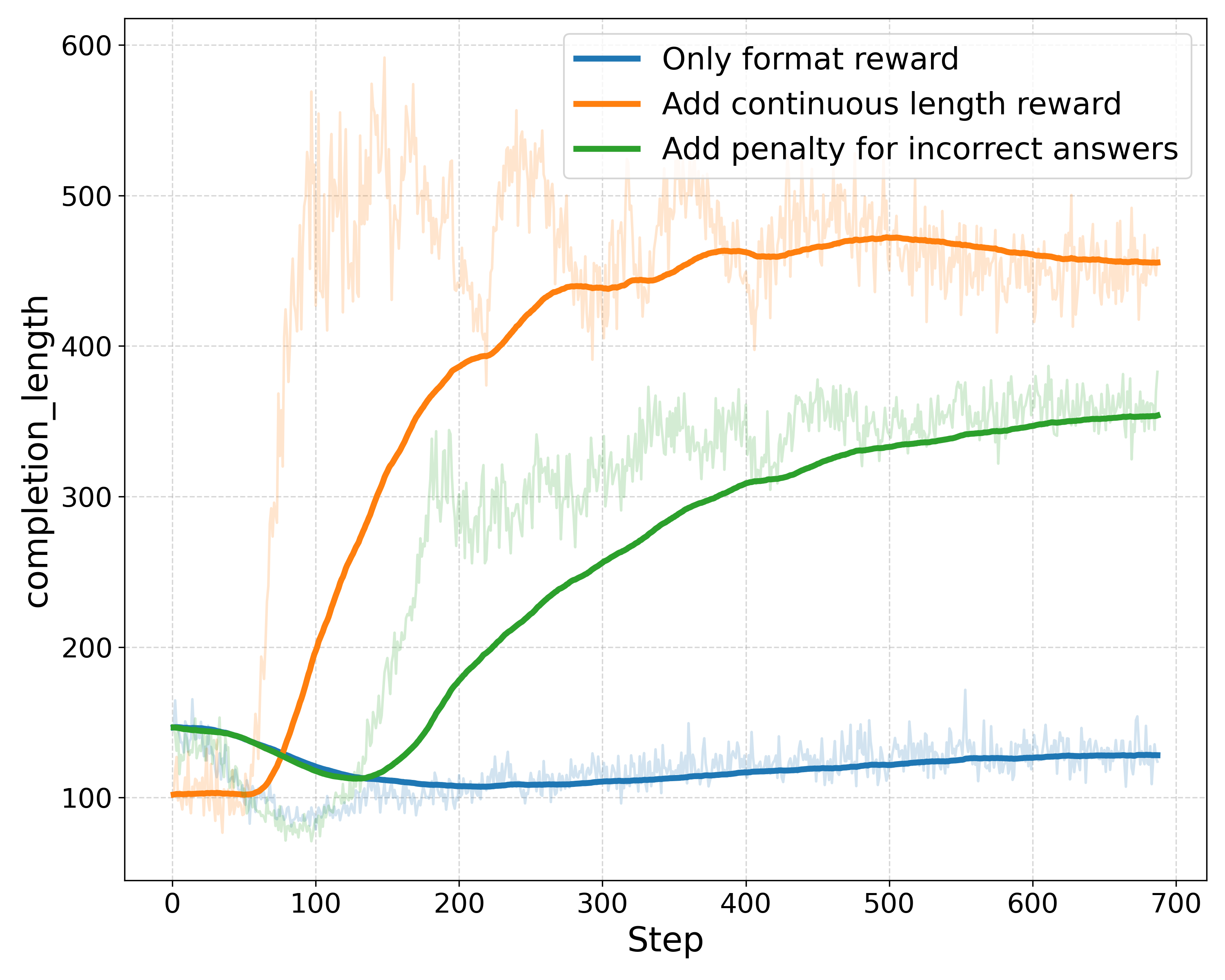}
  \caption{The variation in response length during training under different settings.}
  \label{fig:gline}
  \vspace{-0.1in}
\end{wrapfigure}

In our initial experiments, similar to other works, we only apply format reward without incorporating continuous length reward. However, constrained by the capabilities of small-scale language models, training under this setup does not lead to an increase in response length, and even results in a slight decline. After introducing continuous length reward, the model's response length significantly increases during training, as shown in Figure \ref{fig:gline}. However, we observe that under this setup, the model engages in some meaningless reasoning to increase response length, which does not improve performance and even leads to a significant increase in training time. When incorporating answer correctness penalty into the total reward as described in Section \ref{reward rules}, we observe both qualitative improvements in model responses and continued growth in output length and rewards throughout training as shown in Figure \ref{fig:short}.

\begin{figure*}[h]
  \centering
  \includegraphics[width=\textwidth]{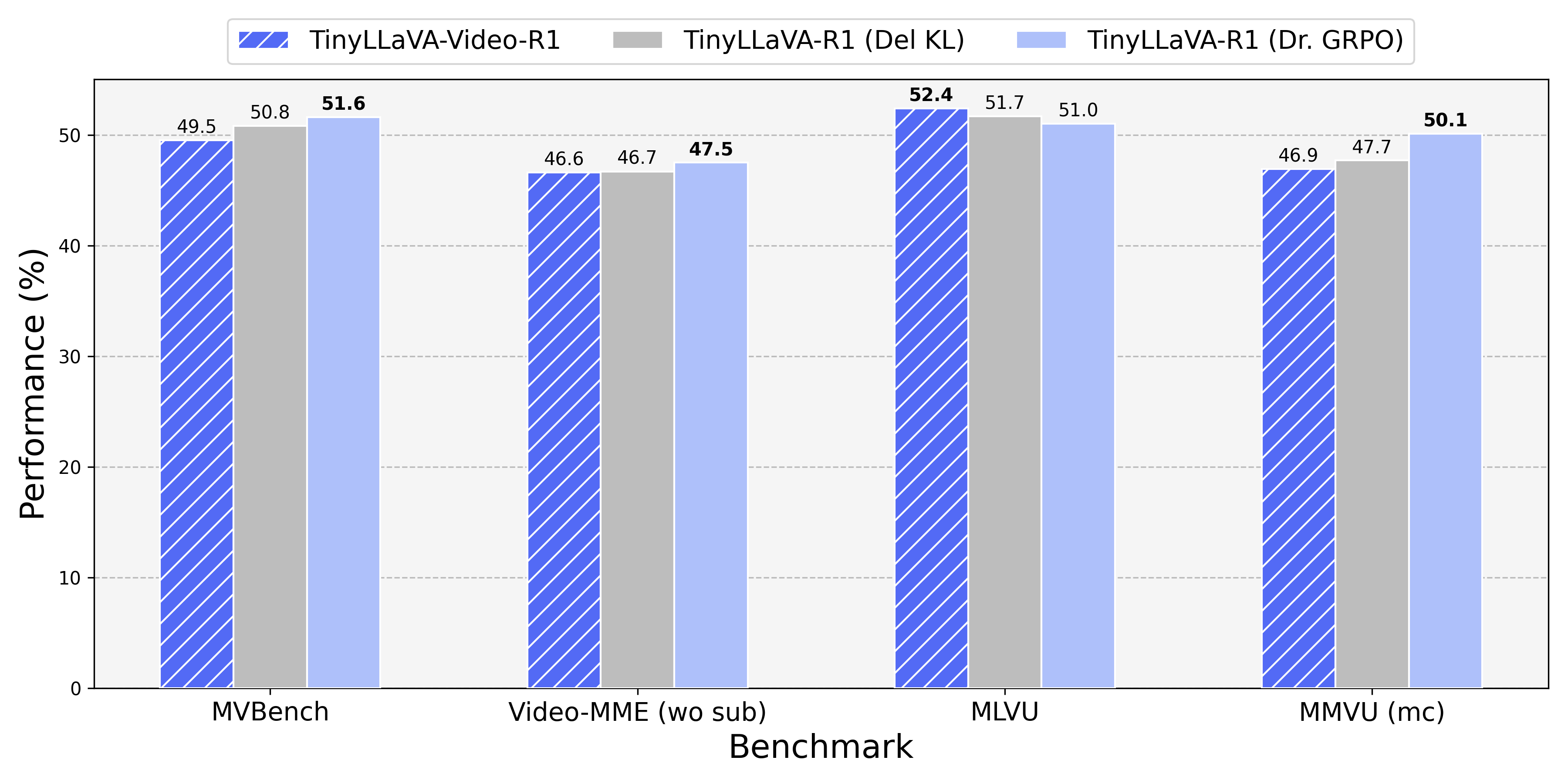}
   \vspace{-0.2in}
   \caption{Ablation study on TinyLLaVA-R1 variants across multiple benchmarks. We compare the original TinyLLaVA-Video-R1 with two ablated versions: removing the KL divergence term (Del KL) and replacing the original GRPO with Dr. GRPO. Results are reported on MVBench, Video-MME (without subtitle input), MLVU, and MMVU (multiple-choice subset). \textbf{Bold} values indicate the best performance for each benchmark. }
   \label{fig:ablation}
   \vspace{-0.1in}
\end{figure*}


\subsubsection{Other Experimental Explorations}

Meanwhile, we also experiment with some existing improvements to GRPO. Some studies \cite{yu2025dapo, liu2025understanding} suggest that the distribution of reasoning models may differ significantly from the initial model, so removing the KL divergence can eliminate constraints on the model. As shown in Figure \ref{fig:ablation}, our experiments similarly demonstrate that eliminating the KL divergence improves model performance. Additionally, Dr. GRPO \cite{liu2025understanding} argues that the increase in response length may also stem from inherent biases in the GRPO objective function. After removing the KL divergence, we further exclude the response length term from the objective function and the reward variance term from the advantage calculation. As shown in Figure \ref{fig:ablation}, the performance of the model improves again. At the same time, we observe a noticeable reduction in response length, the model tends to only provide descriptions of the video content while omitting analysis of the answer. We attribute this to the lack of strong reasoning in the training dataset, which fails to stimulate deep logical reasoning in the models.

\section{Conclusion and Future Work}

In this work, we propose the small-scale video reasoning model TinyLLaVA-Video-R1, which is trained using reinforcement learning on a general Video-QA dataset. It not only significantly enhances reasoning and thinking capabilities, but also exhibits the emergent characteristic of "aha moment". Additionally, we present a series of experimental findings, hoping this work will provide valuable insights for future practitioners exploring the video reasoning abilities of small-scale models. We will further investigate small-scale video reasoning models, with potential future directions as follows:
\begin{itemize}

\item \textbf{Introducing high-quality video reasoning data.} Currently, TinyLLaVA-Video-R1 is trained only on general video question-answering data. We aim to explore the upper limits of the model's reasoning capabilities by introducing higher-quality video reasoning data.

\item \textbf{Improving reinforcement learning algorithms.} Currently, TinyLLaVA-Video-R1 employs the GRPO algorithm for training. However, this approach exhibits notable limitations. To enhance its effectiveness in video reasoning tasks, we plan to refine the algorithm by addressing the key challenges observed in our experiment.
\end{itemize}

\paragraph{Acknowledgment.} 
This work was partially supported by the National Science and Technology Major Project (Grant No. 2022ZD0116310), National Natural Science Foundation of China (Grant No. 62476016), the Fundamental Research Funds for the Central Universities.
{
    \small
    \bibliographystyle{plain}
    \bibliography{main}
}
\clearpage

\end{document}